\documentclass[10pt]{article}
\usepackage{preamble, spconf, amsmath, graphicx}

\usepackage{algorithm}
\usepackage{algpseudocode}
\usepackage{enumitem}
\usepackage{hyperref}
\setlist{nosep, leftmargin=14pt}

\usepackage{mwe} 

\usepackage{xcolor}

\let\origsection\section
\renewcommand\section[1]{\vspace{-2truemm}\origsection{#1}\vspace{-1truemm}}

\let\origsubsection\subsection
\renewcommand\subsection[1]{\vspace{-2truemm}\origsubsection{#1}\vspace{-1truemm}}

\title{SINETRA: a versatile framework for evaluating single neuron tracking in behaving animals}

\name{
  \begin{tabular}{c}Raphael Reme$^{\star\dagger}$\qquad Alasdair Newson$^{\dagger}$\qquad Elsa Angelini$^{\dagger}$\\\qquad
  Jean-Christophe Olivo-Marin$^{\star}$\qquad Thibault Lagache$^{\star}$\end{tabular}
  \vspace{-3truemm}
  }

\address{\small$^{\star}$ Institut Pasteur, Université de Paris-Cité, CNRS UMR 3691, BioImage Analysis Unit F-75015 Paris, France\\
    \small$^{\dagger}$ LTCI, Telecom Paris, Institut Polytechnique de Paris, France\\
    \small Corresponding author: thibault.lagache@pasteur.fr
    }

\usepackage{eso-pic}
\newlength{\xoffset}
\newlength{\xoffsetbis}
\newlength{\yoffset}
\newlength{\yoffsetbis}
\AddToShipoutPicture{
        \AtPageLowerLeft{\put(\LenToUnit{\xoffset},\LenToUnit{\yoffset}){
		    \rotatebox[origin=r]{-90}{\color{gray}{\Large{This work has been submitted to the IEEE for possible publication.}}}}}
        \AtPageLowerLeft{\put(\LenToUnit{\xoffsetbis},\LenToUnit{\yoffsetbis}){
        \rotatebox[origin=r]{-90}{\color{gray}{\Large{Copyright may be transferred without notice, after which this version may no longer be accessible.}}}
}}}

\begin{document}
\ninept
\maketitle
%
\begin{abstract}
Accurately tracking neuronal activity in behaving animals presents significant challenges due to complex motions and background noise.
The lack of annotated datasets limits the evaluation and improvement of such tracking algorithms. To address this, we developed \emph{SINETRA}, a versatile simulator that generates synthetic tracking data for particles on a deformable background, closely mimicking live animal recordings. This simulator produces annotated 2D and 3D videos that reflect the intricate movements seen in behaving animals like \emph{Hydra Vulgaris}. 
We evaluated four state-of-the-art tracking algorithms highlighting the current limitations of these methods in challenging scenarios and paving the way for improved cell tracking techniques in dynamic biological systems.
\end{abstract}
\begin{keywords}
  Single neuron tracking, Simulation, Mechanical forces, Fluorescence noise, Behaving animals
\end{keywords}

\section{Introduction}
\label{sec:intro}
Numerous algorithms have been developed to track neurons in fluorescence live imaging of behaving animals, particularly in the worm \textit{Caenorhabditis (C.) Elegans}, Zebrafish or the freshwater cnidarian \textit{Hydra Vulgaris}. These animals undergo significant body deformations during behaviors such as worm crawling or Hydra's contraction and complex somersaulting \cite{yamamoto2023peptide}. A first class of tracking methods employs analytical approaches, such as Bayesian filtering and global distance minimization, to track neuronal activity \cite{jaqaman2008robust,chenouard2013multiple}. These methods can be complemented with \textit{tracklet} stitching to assemble fragmented cell trajectories into continuous tracks \cite{Lagache2020.06.22.165696emc2,reme2023tracking}. These Bayesian frameworks can also integrate optical flow (OF) estimation of cell velocity allowing for more precise updates of Kalman filters during tracking \cite{remekoft}. A second class of methods relies on deep learning to perform image registration and establish neuron correspondence across different poses of the animal over time. \cite{yu2021fast,park2024automated,ryu2024versatileZephIR}.

To objectively compare cell tracking algorithms, several software tools\cite{svoboda2016mitogen,matyjaszkiewicz2017bsim} and annotated datasets\cite{ulman2017objective} have been developed. However, these mainly address classical cell tracking challenges, focusing on randomly moving and dividing cells \textit{in vitro}, which do not reflect the conditions of neuron tracking in behaving animals. Neurons do not divide and their motion is driven by the animal's body deformations. As a result, evaluating neuron tracking algorithms typically requires labor-intensive manual annotations of time-lapse sequences, which is impractical for large datasets, especially in three dimensions\cite{hanson2024automatic,ryu2024versatileZephIR,Lagache2020.06.22.165696emc2}. 

To facilitate the objective comparison and improvement of SIngle NEurons TRAcking algorithms, we developed \emph{SINETRA}, a versatile framework for simulating two- and three-dimensional time-lapse sequences reproducing the complex motions of neurons in live fluorescence imaging of behaving animals. This simulator models particles (neurons) as spots within a structured background that accounts for the auto-fluorescence of the embedding tissue. The tissue deformations related to animal behaviors are either modeled with a system of damped harmonic oscillators subjected to random contraction or elongation forces, or are directly extracted from optical flow estimates of live animal deformations.

Using this simulation framework, we benchmarked four tracking algorithms on our synthetic data: \emph{eMHT} \cite{chenouard2013multiple} and \emph{u-track} \cite{jaqaman2008robust} that are standard Bayesian filtering and distance minimization methods, \emph{KOFT} \cite{remekoft} that uses optical flow to accurately estimate cell velocity in a Kalman filter and \emph{ZephIR} \cite{ryu2024versatileZephIR} which propagates semi-annotated tracks using image registration. This benchmark underscores the existing limitations of these approaches when dealing with complex movements, opening the door for enhanced cell tracking methods in dynamic biological systems.

\section{Method}

\begin{figure}
    \centering
    \includegraphics[width=\linewidth]{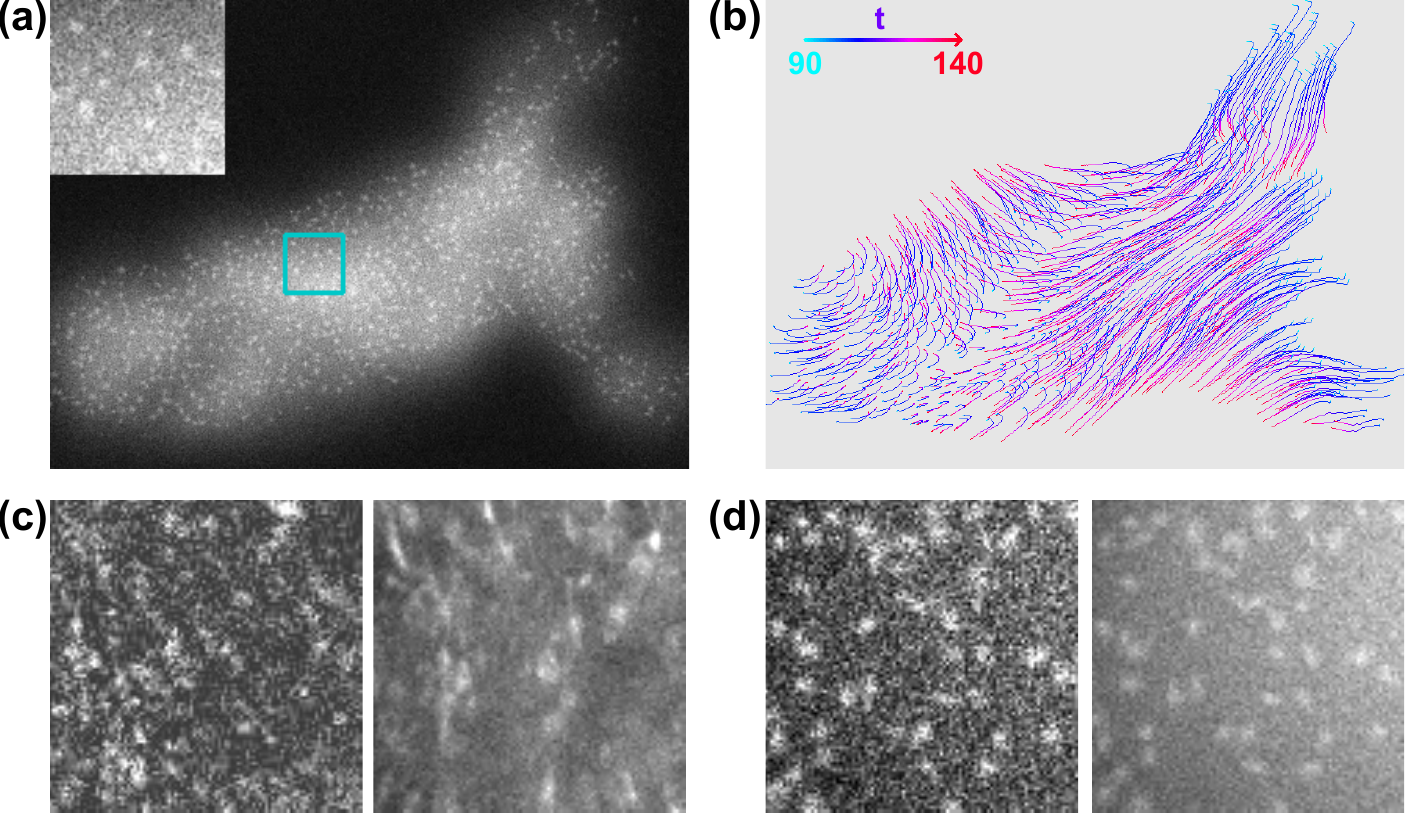}
    \vspace{-4truemm}
    \caption{\ninept\textbf{Synthetic image simulator (Hydra flow) (a)} A synthetic image using the default imaging parameters $\alpha=0.2$ and $\Delta=50$. \textbf{(b)} Synthetic tracks on 50 frames centered around a contraction (frames 90 to 140, with a contraction at $t = 100$). \textbf{(c)} 100x100 pixels from fluorescence videos of \emph{Hydra Vulgaris}' neurons \cite{ hanson2024automatic,dupre2017non}. \textbf{(d)} 100x100 pixels from synthetic videos using different imaging parameters. Left: $\Delta=15$, $\alpha=0.5$. Right: $\Delta=200$, $\alpha=0.2$.}
    \vspace{-4truemm}
  \label{fig:synthetic example}
\end{figure}
\subsection{Modeling fluorescent particles and background}
\subsubsection{Particles}
We first modeled realistic fluorescent particles (neurons). A particle is characterized by its position, intensity and shape, all of which change over time. Formally, each particle $1\leq i\leq N$ in frame $1\leq t \leq T$ is modeled with a $d \in \{2, 3\}$ dimensional Gaussian profile, with $\mathbf{x}_{t,i} \in \R^d$ its position, $w_{t, i} \in \inter{0}{1}$ its intensity weight, and $\mathbf{\Sigma}_{t, i} \in \mathcal{S}_d^{++}$ its elliptic shape, where $\mathcal{S}_d^{++}$ is the ensemble of symmetric and positive matrices in dimension $d$.

On the first frame, we defined the animal's body as the pixels $B\subset \Omega$, where $\Omega = \{\mathbf{z} \in \R^d, \mathbf{z} \ge 0 \text{ and } \mathbf{z} < \mathbf{S}\}$ is the simulation domain and $\mathbf{S}\in \N^d$ the dimensions of the generated images. This pixels $B$ are either obtained from experimental videos using coarse segmentation of the tissue based on intensity thresholding or randomly sampled as an ellipse that covers around 30\% of the surface/volume of the simulation domain $\Omega$. The initial position of particles $\mathbf{x}_{t=0,i}$ are uniformly distributed within $B$ with the sparsity of the distribution tuned via a minimal distance between particles (neurons). Particle intensity weights $w_{t=0, i}$ are initially set to 1 and the axes of elliptic shapes $\mathbf{\Sigma}_{t=0, i}$ are initially sampled with a size from 1 to 3 pixels and randomly rotated (see section \ref{sec:shape}).


Let $\Theta_{t, i} = (w_{t, i}, \mathbf{x}_{t,i}, \mathbf{\Sigma}_{t, i})$ be the parameters of a particle $i$ at frame $t$. The noise-free image of fluorescent particles $I^{\text{p}}_t$ is defined at pixel location $\mathbf{z} \in \Omega$ as:
\begin{equation}
        I^\text{p}_t(\mathbf{z})  = \sum_{i=1}^N q(\mathbf{z};\; \Theta_{t, i})\\
         = \sum_{i=1}^N w_{t, i} e^{-\frac{1}{2} (\mathbf{x}_{t, i} - \mathbf{z})^\text{T}\mathbf{\Sigma}_{t, i}^{-1}(\mathbf{x}_{t, i} - \mathbf{z})}    
\end{equation}
with $q(\mathbf{z};\; \Theta_{t, i})$ the weighted Gaussian probability density function evaluated at pixel $\mathbf{z}$.

Since the intensity weight $w_{t, i}$ are constrained in $\inter{0}{1}$ and the particles are separated by a minimal distance, the brightest pixels in $I_t^p$ will not exceed the maximum gray-level value ($\forall \mathbf{z}, I_t^p(\mathbf{z}) < 1$).

\subsubsection{Background \& Noise}\label{sec:noise}
We model the background fluorescence as multiple Gaussian profiles of parameters $(\Theta^\text{b}_{t, i})_{1 \le i \le N^\text{b}} = (w^\text{b}_{t, i}, \mathbf{x}^\text{b}_{t,i}, \mathbf{\Sigma}^\text{b}_{t, i})_{1 \le i \le N^\text{b}}$. Positions of the background Gaussian profiles are also distributed homogeneously in $B$. The sizes of background profiles are larger than tracked fluorescent particles, ranging from $20$ to $60$ pixels. The noise-free background image $I^\text{b}(\mathbf{z})$ is defined at pixel location $\mathbf{z}$ as:
\begin{equation}
        I^\text{b}_t(\mathbf{z}) = \sum_{i=1}^{N^\text{b}} q(\mathbf{z};\; \Theta^\text{b}_{t, i})
\end{equation}

The background profiles overlap extensively, leading to sums of intensities greater than the weights $w^\text{b}_{t, i}$. Without scaling, the brightest pixels of $I^\text{b}_t$ would exceed the maximum gray-level value. Therefore, we normalize the background images by a fixed constant, which is defined as the maximum value of the \emph{first} background image: $G^\text{b} = \max_\mathbf{z} I^\text{b}_0(\mathbf{z})$.

Particle and background signals are linearly mixed with a proportion $\alpha$. We model the fluctuation of the number of photons detected at each pixel $\mathbf{z} \in \Omega$ using a Poisson Shot Noise process with $\Delta$ the integration time. The noisy image at frame $t$ is generated as:
\begin{equation}
    \begin{aligned}
        \bar{I}_t(\mathbf{z})  &= \alpha I^\text{p}_t(\mathbf{z}) + \frac{(1 - \alpha)}{G^\text{b}} I^\text{b}_t(\mathbf{z}) \\
        I_t(\mathbf{z})  &\sim \frac{1}{\Delta}\mathcal{P}\left(\Delta \bar{I}_t(\mathbf{z})\right)
    \end{aligned}
    \label{eq:image_noise}
\end{equation}

This allows to parametrize the simulator to produce faithful images: $\alpha$ controls the visibility of particles over the background, and $\Delta$ the magnitude of the Poisson shot noise (see Figure \ref{fig:synthetic example}). By default, we use $\alpha=0.20$ and $\Delta=50$ (Figure \ref{fig:synthetic example}.a).

\subsubsection{Intensity variations}
\label{sec:intensity}

Our simulation framework allows to model the dynamics of particle $w_{t, i}$ intensity. This is particularly useful to account for changes in particle fluorescence due to the biophysics of the dye such as calcium indicators in fluorescence imaging of neural activity \cite{nakai2001gcamp,yuste2005fluorescence}. For the sake of simplicity, we have fixed the intensity weights to a constant ($\forall i, j, t,\;w_{t, i} = w^\text{b}_{t, j} = 1$) in all our experiments.

\subsubsection{Shape evolution}
\label{sec:shape}

For each Gaussian profile (particles and background), the covariance matrix of the profile can be separated into size and rotation components:
\begin{equation}
    \mathbf{\Sigma}_{t,i} = \mathbf{R}_{\mathbf{\theta}_{t, i}}^\mathsf T \text{Diag}(\mathbf{\sigma}_{t, i})^2 \mathbf{R}_{\mathbf{\theta}_{t, i}}
\end{equation}
where $\mathbf{\sigma}_{t, i} \in \R_+^d$ are the sizes along each axes of the ellipse, and $\mathbf{R}_{\theta_{t, i}}$ is the rotation matrix of angle $\mathbf{\theta}_{t, i}$. In the 2D case, $\theta_{t, i} \in \R$ is a single angle of rotation around the z-axis. In 3D, $\mathbf{\theta}_{t, i} \in R^3$ are the rotation angles around the x, y and z axes.

We model the smooth random evolution of profiles over time around an equilibrium position with damped harmonic oscillators with random forces (see Section~\ref{sec:oscillators}). The equilibrium value of the rotation angle of a particle $i$ is $\mathbf{\theta}_{\text{eq}, i} = \mathbf{\theta}_{t=0, i} \sim U\left(\inter{0}{\pi}\right)$. Then, to simulate rotation noise, we suppose that random forces are applied to the particles, modeled as i.i.d. Gaussian noise. We choose the magnitude of the force so that $\mathbf{\theta}_{t, i} \sim \mathcal{N}\left(\mathbf{\theta}_{\text{eq}, i}, \frac{\pi}{30}\right)$. We emphasize that $(\mathbf{\theta}_{t, i})_{0\le t \le T}$ are correlated in time and are computed by solving the damped harmonic oscillators equations (not sampled from this Gaussian distribution).

The initial particles sizes $\mathbf{\sigma}_{t=0, i}$ are sampled from a Uniform distribution (1 to 3 pixels for particles, 20 to 60 pixels for background profiles). Intuitively, larger particles should have larger size variations. We therefore define the normalized size as $\mathbf{s}_{t, i} = \frac{\mathbf{\sigma}_{t, i}}{\mathbf{\sigma}_{t=0, i}} \in R_+^d$. Random forces are applied to increase or decrease the normalized size around its equilibrium value of 1. They are modeled as i.i.d. Gaussian noise and chosen so that the  $\mathbf{s}_{t, i} \sim \mathcal{N}(1, 0.05)$.

\subsection{Motion modeling}\label{sec:synthetic_motion}

\begin{figure}
    \centering
    \includegraphics[width=\linewidth]{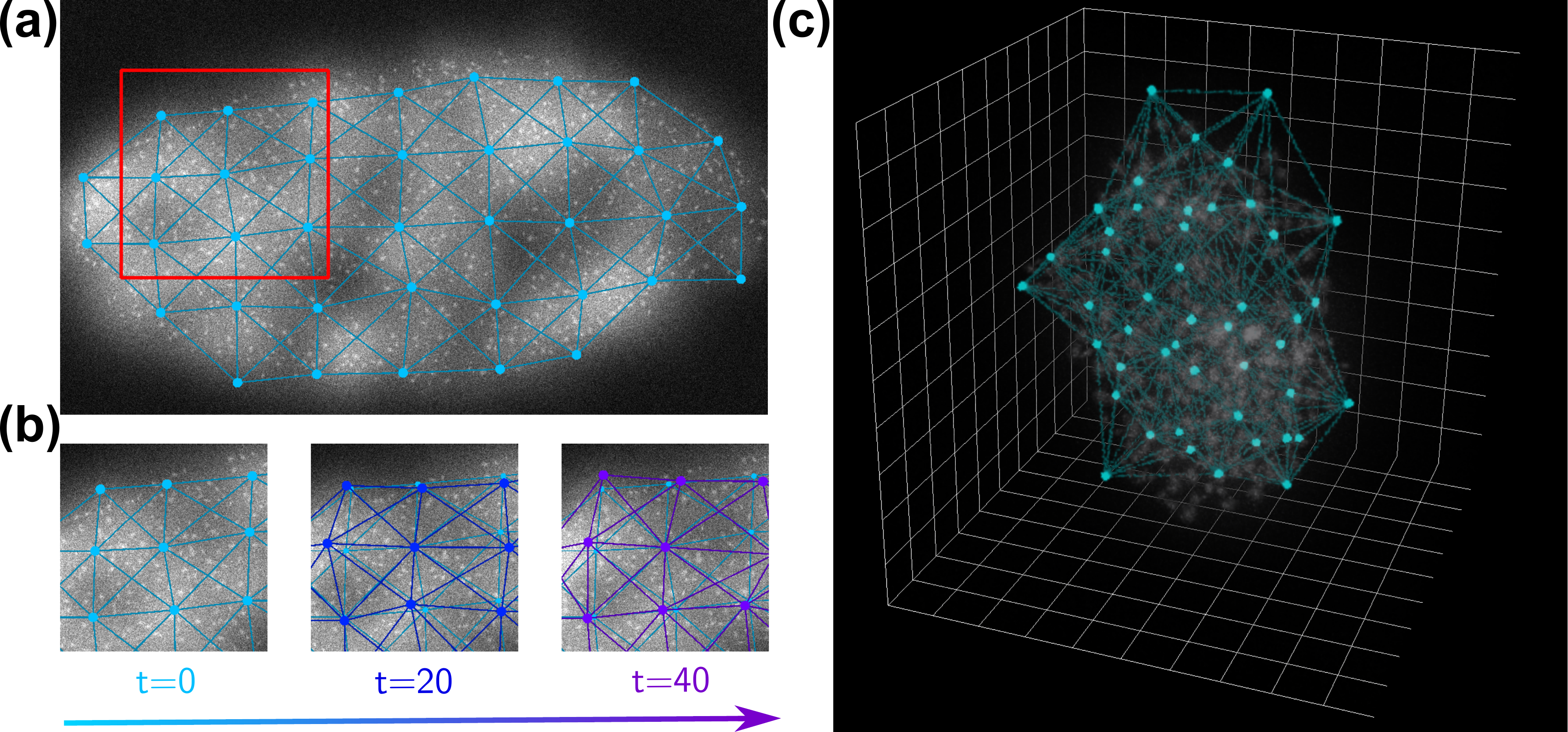}
    \vspace{-4truemm}
    \caption{\ninept\textbf{Synthetic image simulator (Springs 2D/3D). (a)}: Visual appearance of the initial 2D frame with springs between control points (in blue). \textbf{(b)}: Motion induced by random forces and springs constraints over time. The motion of each particle (bright spots) is determined by interpolating control points. \textbf{(c)}: 3D projection of springs (in blue) and noisy particles (in white).}
  \label{fig:springs}
    \vspace{-4truemm}
\end{figure}

To model the motion of neurons embedded in a deformable tissue, we propose two elastic motion models. At each time-step t, we update the position of both particles $\mathbf{x}_{t}$ (neurons) and background profiles $\mathbf{x}^\text{b}_{t}$ using the computed tissue deformation.

\subsubsection{Optical flow motion}
We estimate the animal deformation by computing optical flows between consecutive frames of an experimental video. The optical flows are computed with Farneback algorithm \cite{farneback2003two}, an analytical algorithm that proved to be fast and robust in many applications. The positions of neurons and background profiles are initially sampled within $B$ that corresponds to the thresholded mask of the behaving animals' body, then they are iteratively moved from the computed flow at each time-step (see Figure~\ref{fig:synthetic example}).

\subsubsection{Springs motion}
\label{sec:springs_motion}
To simulate localized, unpredictable deformations of the animal's body, we employ a system of damped springs subjected to random contraction and elongation forces. We position a grid of $n$ control points of equal mass attached by springs within the initial domain $B$ (see Figure \ref{fig:springs}). We then apply random forces to the control points and solve the $n$-body damped harmonic oscillator equations to estimate the dynamics of control points along time (see Section~\ref{sec:oscillators}).

Please note that the control points are not the particles and are not visible on the generated images. The positions of the particles and background Gaussian profiles are computed with an elastic interpolation (Thin Plate Spline) between the $n$ control points.

More formally, let $(\mathbf{p}_i(t) \in \R^d)_{1\le i \le n}$ be the coordinates of the $n$ control points of our system at time $t$. For each point $\mathbf{p}_i$, we define its dampening coefficient $\lambda_i \in \R_+$. For each pair $(\mathbf{p}_i, \mathbf{p}_j)$, a spring is created with a stiffness $k_{ij} \in \R_+$ and an equilibrium length $l^\text{eq}_{ij} \in \R_+$. We set $k_{ij} = 0$ (no spring) for all pairs except the 8 (26 in 3D) closest neighbors. Let $l_{ij}(t)$ be the Euclidean distance between $\mathbf{p}_i$ and $\mathbf{p}_j$ at time $t$. The $n$-body damped harmonic oscillators can be written:
\begin{equation}
    \ddot{\mathbf{p}}_i(t) = \mathbf{f}^\text{springs}_i(t) - \lambda_i \dot{\mathbf{p}}_i(t) + \mathbf{f}^\text{rand}_i(t)
\end{equation}
\begin{equation}
    \mathbf{f}^\text{springs}_i(t) = - \sum_j k_{ij} \left(l_{ij}(t) - l^\text{eq}_{ij}\right)\frac{\mathbf{p}_i(t) - \mathbf{p}_j(t)}{l_{ij}(t)}
\end{equation}
where $\mathbf{f}^\text{rand}_i(t)$ is a random contraction or elongation force applied on control point $i$ at time $t$. To compute this force, we start by randomly selecting a subset $S(t) \subset \Inter{1}{n}$ of control points, where $|S(t)| \sim U\left(\Inter{2}{m}\right)$ and $m=10$ is the fixed maximum number of control points involved in the force. Then, we apply a contraction (resp. elongation) force toward (resp. from) the barycenter of the selected control points. Formally, let $\mathbf{\bar{p}}(t) = \frac{1}{|S(t)|}\sum_{i\in S(t)}\mathbf{p}_i(t)$ be the barycenter of the selected control points. Then:

\begin{equation}
    \mathbf{f}^\text{rand}_i(t) = \vast\{\begin{aligned}& 0 & \text{if } i \notin S(t)\\
            &d(t)a_i(t)\frac{\mathbf{p}_i(t) - \mathbf{\bar{p}}(t)}{\norm{\mathbf{p}_i(t) - \mathbf{\bar{p}}(t)}_2} & \text{otherwise}
    \end{aligned}
\end{equation}
where $d(t) \in \{-1, 1\}$ controls the random direction of the motion (contraction or elongation) and $a_i(t) \sim U(\frac{1}{2}a_\text{max}, a_\text{max})$ is the random amplitude of the motion.

\subsubsection{Damped harmonic oscillators}
\label{sec:oscillators}
In our simulator, we rely on damped harmonic oscillators to model the temporal evolution of positions, angles and sizes of particles. We detail here the calculations underlying these equations in the one-dimensional case with a single spring, which benefits from theoretical background. For our multiple interacting springs (Section \ref{sec:springs_motion}), we have extrapolated the results from this simpler case.

Let $x(t) \in R$ be the quantity of interest at time $t$, we have the following system:

\begin{equation}
    \ddot{x}(t) = f(t) - \lambda \dot{x}(t) - k\left(x(t) - x_{eq}\right)
\end{equation}
where $f(t)$ is a random force, $\lambda$ the dampening coefficient, $k$ the stiffness coefficient and $x_{eq}$ the equilibrium state of the system. The mass of the system is canceled out for simplicity.

The system is discretized with a time interval $dt$: $x_n = x(ndt)$. We decided to use the semi-implicit Euler method which is a symplectic integration method suited for our Hamiltonian system. Starting from initial conditions $x_0, \dot{x}_0$, it computes the next state following:
\begin{equation}
    \begin{aligned}
        \ddot{x}_{n} &= f_{n} - \lambda \dot{x}_n - k (x_n - x_{eq})\\
        \dot{x}_{n+1} &= \dot{x}_n + dt\ddot{x}_{n}\\
        x_{n+1} &= x_n + dt\dot{x}_{n+1}
    \end{aligned}
\end{equation}

For the processes modeled here, oscillations are not realistic. Therefore we decided to critically damp the system so that the transient solution (without random forces) decays to the equilibrium state. This implies to set $\lambda = 2\sqrt{k}$.

Finally, we introduce the critical time $\tau = \frac{2}{\lambda}$ which corresponds to the exponential decay of the transient solution. We decided to use a reasonable value of $\tau = 10$ frames: without forces, the system returns to the equilibrium in around 10 frames. In all our springs, we have $\lambda = \frac{2}{\tau}$ and $k = \frac{1}{\tau^2}.$

\section{Results and Discussions}

\subsection{Evaluation}
Using our simulator, we generated a synthetic dataset of fluorescent neurons in a behaving animal. We simulated 800 neurons over 200 frames, using $\Delta = 50$ and $\alpha = 0.2$. We focused on three different scenarios:
\begin{enumerate}
    \item \textbf{Hydra Flow}: we generated 1024x1024 images using optical-flow based motion. The flows are extracted from 200 selected frames of an experimental video where the \emph{Hydra Vulgaris} is contracting \cite{dupre2017non}.
    \item \textbf{Springs 2D}: we used springs-based motion (with the amplitude of random forces $a_\text{max} = 4$ pixels) to generate 1024x1024 images.
    \item \textbf{Springs 3D}: we simulated 200x200x200 volumes with springs-based motion (with $a_\text{max} = 3$ pixels).
\end{enumerate}
For a set of ground-truth and computed tracks, we measure the accuracy of the tracking algorithm with the HOTA score \cite{luiten2021hota} that estimates both localization, detection and association performance. We used a tolerance distance $\eta = 2$ pixels (\textit{i.e.} a predicted particle is never associated with a ground-truth particle distant by more than $\eta$ pixels). 

The simulator and the tracking experiments are available alongside these synthetic datasets at \url{https://github.com/raphaelreme/SINETRA}. 

\subsection{Evaluated algorithms}
Using this synthetic dataset, we compared four state-of-the-art tracking algorithms that are based on distinct methodological frameworks: \textbf{(1)} \emph{u-track} \cite{jaqaman2008robust} from trackmate/Fiji software \cite{schindelin2012fiji, tinevez2017trackmate} is a global distance minimization algorithm in two steps (frame-to-frame linking, tracklet stitching). We used the advanced version that model motion with Kalman filters. \textbf{(2)} \emph{eMHT} \cite{chenouard2013multiple} from Icy software \cite{de2012icy} is a probabilistic algorithm based on multiple motion models, with a probabilistic handling of tracks and a multiple hypothesis association method. 
\textbf{(3)} \emph{KOFT} \cite{remekoft} implemented in ByoTrack \cite{hanson2024automatic} is a probabilistic algorithm that exploits optical flow within Kalman filters. \textbf{(4)} \emph{ZephIR} \cite{ryu2024versatileZephIR} is a frame by frame registration algorithm that propagates partially annotated tracks. We tested \emph{ZephIR} with three (\emph{ZephIR@3}) and ten (\emph{ZephIR@10}) annotated frames.

Each algorithm requires a pre-detection of the particles to track. We chose to rely on the wavelet thresholding method described in \cite{OLIVOMARIN20021989} to detect particles for which we measured a detection f1 score of around 80\% across our synthetic dataset. 

\subsection{Results}

\begin{table}
\ninept
  \begin{center}
    \begin{tabular}{|l|c|c|c|}
      \hline
      Scenario         & Hydra Flow                 & Springs 2D             & Springs 3D               \\
      \hline
      \emph{u-track}   & $67.7 \pm 5.7\%$ & $80.3 \pm 3.7\%$ & $70.8 \pm 6.2\%$ \\
      \emph{eMHT}      & $73.4 \pm 8.1\%$ & $82.4 \pm 2.8\%$ & $76.5 \pm 6.4\%$ \\
      \emph{KOFT}      & $\mathbf{93.5} \pm 2.1\%$ & $\mathbf{95.8} \pm 0.6\%$ & $\mathbf{80.2} \pm 5.8\%$ \\
      \emph{ZephIR@3}  & $76.4 \pm 6.0\%$ & $74.6 \pm 7.4\%$ & $51.5 \pm 2.5\%$ \\
      \emph{ZephIR@10} & $91.9 \pm 3.2\%$ & $91.3 \pm 1.7\%$ & $75.3 \pm 2.7\%$ \\
      \hline
    \end{tabular}
  \end{center}
  \vspace{-4truemm}
  \caption{HOTA at 2 pixels \cite{luiten2021hota} of the different tracking algorithms, on three synthetic scenarios. We report the mean and std on 5 different random simulations sharing the same parameters (except the random seed).}
  \label{tab:results}
  \vspace{-2truemm}
\end{table}

\begin{figure}
    \ninept 
    \centering
    \includegraphics[width=\linewidth]{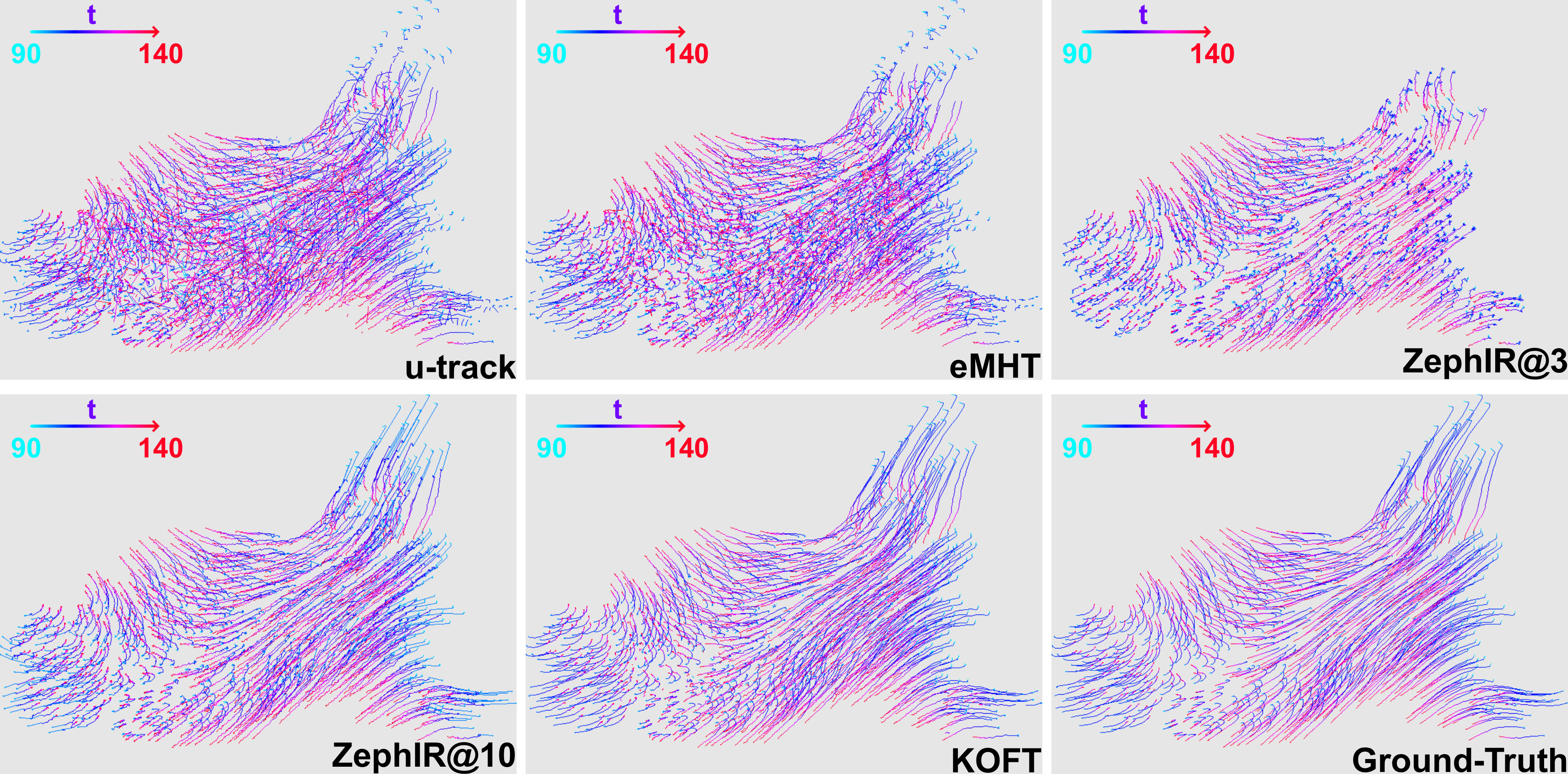}
    \vspace{-4truemm}
    \caption{Temporal projection of tracks (Hydra flow scenario) on 50 frames centered around a contraction (frames $90$ to $140$, contraction at $t=100$).}
    \label{fig:results}
    \vspace{-4truemm}
\end{figure}

Table \ref{tab:results} summarizes the performance of the different tracking algorithms on the three simulation scenarios.  \emph{u-track} \cite{jaqaman2008robust, schindelin2012fiji, tinevez2017trackmate} 
and \emph{eMHT} \cite{chenouard2013multiple, de2012icy} model a near-constant velocity of tracked particles, and are less robust to large and sudden contractions that occurs in behaving animals compared to \emph{KOFT} and \emph{ZephIR} (see Figure \ref{fig:results}). \emph{u-track} reaches an averaged of $67.7\%$ HOTA on the Hydra flow scenario, whereas \emph{eMHT} is slighlty more robust thanks to its multiple hypothesis association and reaches $73.4\%$ HOTA.

\emph{ZephIR} registration method is more resilient to the animal's large motions thanks to well chosen annotated frames \cite{ryu2024versatileZephIR}. Using 3 annotated frames in elongated and contracted animals, it outperforms \emph{u-track} and \emph{eMHT} on Hydra Flow scenario, reaching 80.9\% HOTA. Yet, on springs-based videos where contractions are more frequent, \emph{ZephIR} requires more annotated frames to outperforms \emph{u-track} and \emph{eMHT}. We emphasize that frame annotating is tedious in experimental datasets, limiting the capabilities of \emph{ZephIR} in highly deforming organisms like Hydra. Finally, \emph{KOFT} \cite{remekoft, hanson2024automatic} outperformed the other methods across the different simulations, due to its precise estimation of neuronal flow motion and the use of Kalman filters, both of which significantly improve particle tracking. (see Figure \ref{fig:results}). It reaches above 90\% HOTA on 2D scenarios. However, we noticed that it had difficulty tracking neurons during the animal's fastest movements when neurons were densely packed, or in 3D where its optical flow estimates are less precise (80.2 \% on springs 3D scenario).

We also compared the computational efficiencies of the different tracking methods. \emph{u-track} is the fastest method, running at 25 FPS on a Lenovo Legion 5 laptop on our 2D/3D sequences. The multiple hypothesis association in \emph{eMHT} is more expensive, and the algorithm runs at 2 FPS on 2D/3D sequences. \emph{KOFT} is slower in 3D than in 2D because of the optical flow computations, running at 10 FPS in 2D and 2 FPS in 3D. Finally \emph{ZephIR} is the slowest option because it needs to optimize its loss at each frame. It runs at 0.3 FPS on CPU but is much faster on GPU, running at 3 FPS on a nvidia RTX3070.

\section{Conclusion}

In this paper, we presented \emph{SINETRA}, a versatile framework to generate synthetic annotated single-particle-tracking datasets with a realistic imaging noise and complex motions such as those experienced by neurons in behaving animals like \emph{Hydra Vulgaris} or \emph{C. elegans}. To model animal's deformation and neurons' motion, we either use optical flow estimates from experimental datasets, or model elastic deformations with a system of damped springs that experience random contraction and elongation forces.

Through various simulation scenarios in both 2D or 3D, we assessed and highlighted the limitations of four different tracking algorithms. We believe that developing realistic simulation frameworks is crucial for creating more robust tracking algorithms for \textit{in vivo} monitoring of neuronal activity in behaving animals.

\section{Compliance with ethical standards}
This is a numerical simulation study for which no ethical approval was required.

\section{Acknowledgments}
This work is supported by the Institut Pasteur and France-BioImaging Infrastructure (ANR-10-INBS-04). R.R and T.L. are supported by the ANR (ANR-21-CE45-0020-01 REBIRTH).

None of the authors declare to have a financial conflict of interest in the results of this study.

\bibliographystyle{IEEEbib_short}
\bibliography{refs}
\vfill
\pagebreak

\end{document}